\documentclass{article}

\usepackage{microtype}
\usepackage{graphicx}
\usepackage{subfigure}
\usepackage{booktabs} 
\usepackage{relsize}

\usepackage{hyperref}

\usepackage{nicefrac}       
\usepackage{url}
\usepackage{todonotes}

\newcommand{\tablefont}{\fontfamily{cmss}\fontseries{uc}\fontsize{10pt}{9pt}\selectfont\centering}
\newcommand{\tablefontsmaller}{\fontfamily{cmss}\fontseries{uc}\fontsize{8pt}{7pt}\selectfont\centering}

\usepackage{amsbsy}
\usepackage{amsmath}
\usepackage{amsmath}%
\usepackage{ntheorem}       
\usepackage{dsfont}
\usepackage{MnSymbol}%
\usepackage{wasysym}%
\usepackage[cal=dutchcal,
calscaled=1,
bb=boondox,
scr=euler]{mathalfa}
\usepackage{wrapfig}
\usepackage[capitalise]{cleveref}
\crefformat{equation}{(#2#1#3)}

\newcommand{\ie}[0]{\emph{i.e.},~}
\newcommand{\eg}[0]{\emph{e.g.},~}

\newcommand{\mat}[1]{\ensuremath{{#1}}}
\newcommand{\ident}[0]{\ensuremath{\mathbf{1}}}
\newcommand{\gr}[1]{\ensuremath{\mathcal{#1}}}
\newcommand{\set}[1]{\ensuremath{\mathbb{#1}}}
\renewcommand{\vec}[1]{\ensuremath{\operatorname{vec}({#1})}}
\newcommand{\unvec}[1]{\ensuremath{\operatorname{vec}^{-1}({#1})}}

\newcommand{\pN}[0]{\ensuremath{\mathscr{N}}}

\newcommand{\bn}[0]{\ensuremath{\underline{n}}}
\newcommand{\bm}[0]{\ensuremath{\underline{m}}}
\newcommand{\bi}[0]{\ensuremath{\underline{k}}}
\newcommand{\st}{\;:\;}
\newcommand{\prm}[1]{\ensuremath{^{(#1)}}}
\newcommand{\chn}[1]{\ensuremath{^{\langle#1 \rangle}}}

\newcommand{\grn}[2]{\ensuremath{\gr{#1}\prm{#2}}}
\newcommand{\Trp}[0]{\ensuremath{^{\mathsf{T}}}}

\newcommand{\XX}[0]{\ensuremath{\mat{X}}}
\newcommand{\YY}[0]{\ensuremath{\mat{Y}}}
\newcommand{\WW}[0]{\ensuremath{\mat{W}}}
\newcommand{\ZZ}[0]{\ensuremath{\mat{Z}}}
\newcommand{\xx}[0]{\ensuremath{\mat{x}}}

\renewcommand{\Re}[0]{\ensuremath{\mathds{R}}}

\newtheorem{theorem}{Theorem}[section]

\newtheorem{proposition}[theorem]{Proposition}

\theoremstyle{definition}
\newtheorem{definition}{Definition}[section]

\theoremsymbol{\ensuremath{\blacksquare}}
\newtheorem*{proof}{Proof}

\newcommand{\row}[0]{\ensuremath{\set{R}}}
\newcommand{\col}[0]{\ensuremath{\set{C}}}
\newcommand{\ind}[0]{\ensuremath{\set{I}}}

\usepackage[accepted]{icml2018}

\begin{document}

\twocolumn[
\icmltitle{Deep Models of Interactions Across Sets}



\icmlsetsymbol{equal}{*}

\begin{icmlauthorlist}
\icmlauthor{Jason Hartford}{equal,ubc}
\icmlauthor{Devon R Graham}{equal,ubc}
\icmlauthor{Kevin Leyton-Brown}{ubc}
\icmlauthor{Siamak Ravanbakhsh}{ubc}
\end{icmlauthorlist}

\icmlaffiliation{ubc}{Department of Computer Science, University of British Columbia, Canada}

\icmlcorrespondingauthor{Devon Graham}{drgraham@cs.ubc.ca}
\icmlcorrespondingauthor{Jason Hartford}{jasonhar@cs.ubc.ca}

\icmlkeywords{deep learning, exchangeability, equivariance, invariance, tensor, matrix completion, recommender systems}

\vskip 0.3in
]



\printAffiliationsAndNotice{\icmlEqualContribution} 

\begin{abstract}
    We use deep learning to 
    model interactions across two or more sets of objects, such as user--movie ratings, protein--drug bindings, or ternary user-item-tag interactions. 
    The canonical representation of such interactions is a matrix (or a higher-dimensional tensor) with an exchangeability property: the encoding's meaning is not changed by permuting rows or columns. We argue that models should hence be \emph{Permutation Equivariant (PE)}: constrained to make the same predictions across such permutations. We present a parameter-sharing scheme and prove that it could not be made any more expressive without violating PE.
This scheme yields three benefits. First, we demonstrate state-of-the-art performance on multiple matrix completion benchmarks. Second, our models require a number of parameters independent of the numbers of objects, and thus scale well to large datasets. Third, models can be queried about new objects that were not available at training time, but for which interactions have since been observed. In experiments, our models achieved surprisingly good generalization performance on this \emph{matrix extrapolation} task, both within domains (e.g., new users and new movies drawn from the same distribution used for training) and even across domains (e.g., predicting music ratings after training on movies).
\end{abstract}


\section{Introduction}\label{sec:intro}

Suppose we are given a set of users $\set{N} = \{1,\ldots, N\}$, a set of movies $\set{M} = \{1,\ldots,M\}$, and their interaction in the form of tuples $\set{X} = \langle n, m, x\rangle$ with $n \in \set{N}$, $m \in \set{M}$ and $x \in \mathds{R}$. Our goal is to learn a function that models the interaction between users and movies, i.e., mapping from $N \times M$ to $\Re$. The canonical representation of such a function is a matrix $\XX \in \Re^{N \times M}$; of course, we want $\XX_{n,m} = x$ for each $\langle n, m, x\rangle \in \set{X}$. Learning our function corresponds to \emph{matrix completion}: using patterns in $\set{X}$ to predict values for the remaining elements of $\XX$.
$X$ is what we will call an \emph{exchangeable matrix}: any row- and column-wise permutation of $X$ represents the same set of ratings and hence the same matrix completion problem. 

{Exchangeability}  has a long history in  machine learning and statistics.
{de Finetti}'s theorem states that exchangeable sequences are the product of a {latent variable model}. Extensions of this theorem characterize distributions over other exchangeable structures, from matrices to graphs; see \citet{orbanz2015bayesian} for a detailed treatment.
In machine learning, a variety of frameworks formalize exchangeability in data, from plate notation to {statistical relational models}~\citep{raedt2016statistical,getoor2007introduction}. 
When dealing with exchangeable arrays (or tensors), a common approach is {tensor factorization}. In particular, one thread of work leverages tensor decomposition for inference in latent variable models~\citep{anandkumar2014tensor}.
However, in addition to having limited expressive power, tensor factorization requires retraining models for each new input. 


We call a learning algorithm \emph{Permutation Equivariant (PE)} if it is constrained to give the same answer across all exchangeable inputs; we argue that PE is an important form of inductive bias in exchangeable domains. However, it is not trivial to achieve; indeed, all existing parametric factorization and matrix/tensor completion methods associate parameters with each row and column, and thus are not PE. How can we enforce this property? One approach is to augment the input with all $M! \times N!$ permutations of rows and columns.
However, this is computationally wasteful and becomes infeasible for all but the smallest $N$ and $M$. 
Instead, we show that a simple {parameter-sharing} scheme suffices to ensure that a deep model can represent only PE functions. 
The result is analogous to the idea of a convolution layer: a lower-dimensional effective parameter space that enforces a desired equivariance property. 
Indeed, parameter-sharing is a generic and efficient approach for achieving equivariance in deep models~\citep{ravanbakhsh_symmetry}. 




\begin{figure}[t]\centering
\includegraphics[width=\columnwidth]{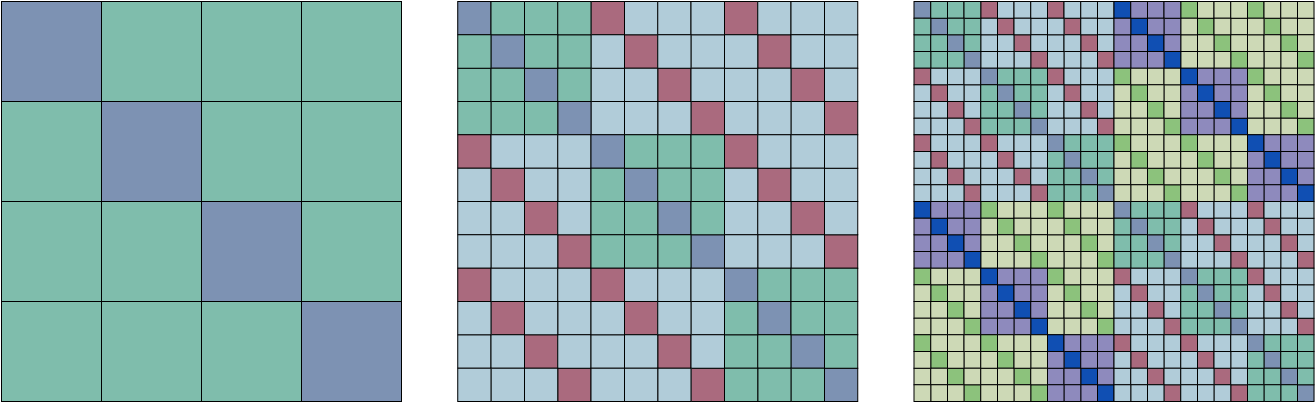}
\caption{Structure of our parameter matrix for the 1D (left), 2D (centre), and 3D (right) input arrays. 
The parameter-sharing patterns for the weight matrix of the higher dimensional arrays can be produced via the \textit{Kronecker product} of the weight matrix for the 1D array (\ie vector).}
\label{fig:parameter_matrix}
\end{figure}

%

When the matrix models the interaction between the members of the same group, one could further constrain permutations to be identical across both rows and columns. 
An example of such a \textit{jointly exchangeable matrix} \mbox{\citep{orbanz2015bayesian}}, 
modelling the interaction of the nodes in a graph, is the adjacency matrix deployed by graph convolution. Our approach reduces to graph convolution in the special case of 2D arrays with this additional parameter-sharing constraint, but also applies to arbitrary matrices and higher dimensional arrays.



Finally, we explain connections to some of our own past work. First, we introduced a similar parameter-sharing scheme in the context of behavioral game theory \citep{hartford2016deep}: rows and columns represented players' actions and the exchangeable matrix encoded payoffs. The current work provides a theoretical foundation for these ideas and shows how a similar architecture can be applied much more generally. Second, our model for exchangeable matrices can be seen as a generalization of the \emph{deep sets} architecture \citep{zaheer_deepsets}, where a set can be seen as a one-dimensional exchangeable array. 

In what follows, we begin by introducing our parameter-sharing approach in \cref{sec:layer}, considering the cases of both dense and sparse matrices.
In \cref{sec:architectures} we study two architectures for matrix completion that use an exchangeable matrix layer. In particular
the factorized autoencoding model provides a powerful alternative to commonly used matrix factorization methods; notably, it does not require retraining to be evaluated on previously unseen data.
In \cref{sec:empirical} we present extensive results on benchmark matrix completion datasets. We generalize our approach to higher-dimensional tensors in \cref{sec:tensors}. 

\section{Exchangeable Matrix Layer}\label{sec:layer}
\begin{figure}[t]\centering
\includegraphics[width=\columnwidth]{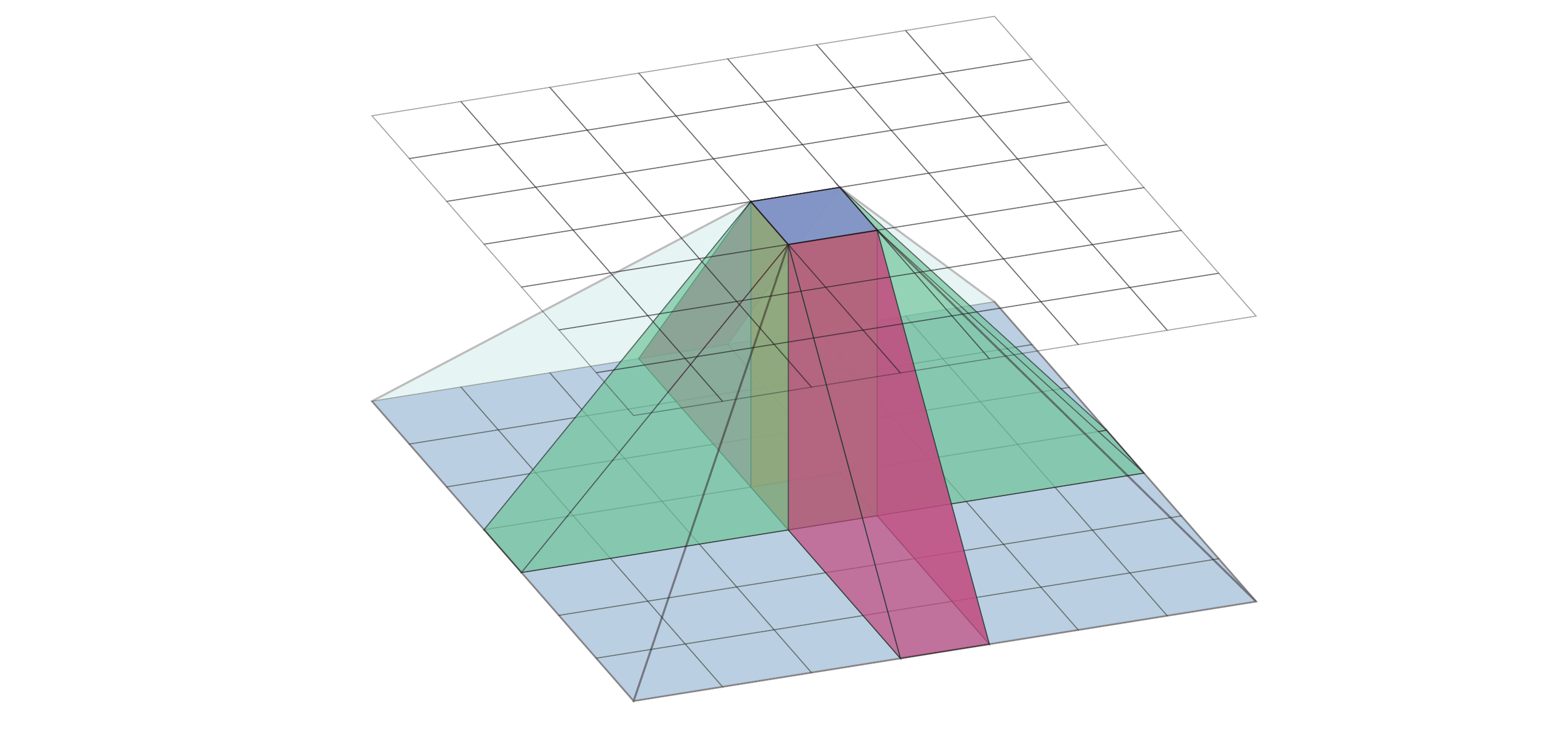}
\vspace{-2em}\caption{Parameter sharing in an exchangeable matrix layer. 
The application of different tied parameters to input elements is illustrated using dark blue for $w_1$, green for  $w_2$, red for $w_3$, and light blue for $w_4$. The same structure (with the same four parameters) repeats across all units of the output layer. }
\label{fig:parameter_sharing}
\end{figure}

Let  $\XX \in \Re^{N \times M}$ be our ``exchangeable'' input matrix. We use $\vec{\XX} \in \Re^{N M}$ to denote its vectorized form and $\unvec{\xx}$ to denote the inverse of the vectorization that reshapes a vector of length $N M$ into an $N \times M$ matrix---\ie $\unvec{\vec{\XX}} = \XX$.
Consider a fully connected layer ${\YY} := \unvec{\sigma(W \vec{\XX})}$ where $\sigma$ is an element-wise nonlinear function such as sigmoid, $W \in \Re^{N M \times N M}$,
and ${Y} \in \Re^{N \times M}$ is the output matrix. 
Exchangeablity of $\XX$ motivates the following property:
suppose we permute the rows and columns $\XX$
using two arbitrary permutation matrices $G\prm{N} \in \{0,1\}^{N \times N}$ and $G\prm{M} \in \{0,1\}^{M \times M}$
to get $\XX' := G\prm{N} \XX G\prm{M}$. Permutation Equivariance (PE) requires the new output matrix $\YY' := \unvec{\sigma(W \vec{X'})}$ to 
experience the same permutation of rows and columns---that is, we require $\YY' = G\prm{N} \YY G\prm{M}$.

\begin{definition}[exchangeable matrix layer, simplified\footnotemark]\footnotetext{This definition is simplified to ease exposition; the full definition (see \cref{sec:tensors}) adds the additional constraint that the layer \emph{not} be equivariant wrt any other permutation of the elements of $\XX$. Otherwise, a trivial layer with a constant weight matrix $W_{n,m} = c$ would also satisfy the stated equivariance property.}\label{def:2pe_layer}

Given $X \in \Re^{N \times M}$, a fully connected layer 
$\sigma(W \vec{X})$ with $W \in \Re^{NM \times NM}$ is called an {exchangeable matrix layer} if, for all permutation matrices $G\prm{N} \in \{0,1\}^{N \times N}$ and $G\prm{M} \in \{0,1\}^{M \times M}$, permutation of the rows and columns results in the same permutations of the output:
\begin{align}\label{eq:1}
& \unvec{\sigma(W \vec{G\prm{N} \XX G\prm{M}})} = \\ \nonumber
& \hspace{2cm} G\prm{N} \unvec{\sigma(W \vec{\XX})} G\prm{M}.
\end{align} 
\end{definition}

This requirement heavily constrains the weight matrix $W$: indeed, its number of effective degrees of freedom cannot even grow with $N$ and $M$. Instead, the resulting layer is forced to have the following, simple form:
\begin{align}\label{eq:w}
  \WW_{(n, m), (n',m')} :=
  \begin{cases}
    w_1 & n = n', m = m'\\
    w_2 & n = n', m \neq m' \\
    w_3 & n \neq n', m = m' \\
    w_4 & n \neq n', m \neq m' . \\
  \end{cases}
\end{align}

For each output element $\YY_{n,m}$, we have the following parameters: one connecting it to its counterpart $\XX_{n,m}$; 
one each connecting it to the inputs of the same row and the inputs of the same column; and one shared by all the other connections. We also include a bias parameter; see \cref{fig:parameter_sharing} for an illustration of the action of this parameter matrix, and \cref{fig:parameter_matrix} for a visual illustration of it. Theorem \ref{thm:pe_layer} formalizes the requirement on our parameter matrix. All proofs are deferred to the appendix. 

\begin{theorem}\label{thm:pe_layer}
Given a strictly monotonic function $\sigma$, a neural layer $\sigma(W \vec{X})$ is an \textbf{exchangeable matrix layer} \textit{iff} the elements of the parameter matrix $W$ are tied together such that the resulting fully connected layer simplifies to 
\begin{align}
  \label{eq:layer_simple_mat}
  \YY = \sigma & \left ( w_1' \XX + \frac{w_2'}{N}(\ident_N\ident_N\Trp \XX) + \frac{w_3'}{M}(\XX\ident_M\ident_M\Trp) \right. \\ \nonumber
&+ \left. \frac{w_4'}{N M}(\ident_N\ident_N\Trp\XX\ident_M\ident_M\Trp) + w_5'\ident_N\ident_M\Trp \right) 
\end{align}
where $\ident_{N} = \underbrace{[1,\ldots,1]\Trp}_{\text{length} N}$ and $w_1',\ldots,w_5' \in \Re$.
\end{theorem}
\vspace{-.6em minus .3em}
This parameter sharing is translated to summing or averaging elements across rows and columns; more generally, PE is preserved by any commutative pooling operation.
Moreover, stacking multiple layers with the same equivariance properties
preserves equivariance~\cite{ravanbakhsh_symmetry}. This allows us to build \textit{deep} PE models.

\paragraph{{Multiple Input--Output Channels}}
Equation~\cref{eq:layer_simple_mat} describes the layer as though it has single input and output matrices.
However, as with convolution, we may have $K$ input and $O$ output channels. We use superscripts $\XX\chn{k}$ and $\YY\chn{o}$
to denote such channels. Cross-channel interactions are fully connected---that is, we have five unique parameters $w_1\chn{k,o},\ldots,w_4\chn{k,o}$ for \textit{each combination}
of input--output channels; note that the bias parameter $w_5$ does not depend on the input.
Similar to convolution, the number of channels provides a tuning nob for the expressive 
power of the model.
In this setting, the scalar output $\YY_{n,m}\chn{o}$ is given as
\begin{align}
  \label{eq:layer_multi_channel_mat}
  \YY\chn{o}_{n,m} &= \sigma  \left ( \sum_{k=1}^{K} \bigg ( w_1\chn{k, o} \XX_{n,m}\chn{k} +  \frac{w_2\chn{k,o}}{N}(\sum_{n'} \XX_{n',m}\chn{k}) + \right. \\ \nonumber
    & \left.  \frac{w_3\chn{k,o}}{M}(\sum_{m'} \XX_{n,m'}\chn{k}) +  \frac{w_4\chn{k,o}}{N M}(\sum_{n',m'} \XX_{n',m'}\chn{k}) + w_5\chn{o}\bigg) \right) 
\end{align} 


\vspace{-.6em minus .3em}\paragraph{{Input Features for Rows and Columns}} 
In some applications, in addition to the matrix $\XX \in \Re^{N \times M \times K}$, where $K$ is the number of input channels/features, we may have additional features for rows $R \in \Re^{N \times K'}$ and/or columns $C \in \Re^{M \times K''}$. We can preserve PE by broadcasting these row/column features over the whole matrix and treating them as additional input channels.

\vspace{-.6em minus .3em}\paragraph{{Jointly Exchangeable Matrices}}
For jointly exchangeable matrices, such as adjacency matrix, Equation~\cref{eq:1} is constrained to have $N = M$ and $G\prm{N} = G\prm{M}$.
This will in turn constrain the corresponding parameter-sharing so that $w_2 = w_3$ in Equation~\cref{eq:w}.

\subsection{Sparse Inputs}\label{sec:sparse}
Real-world arrays are often extremely sparse. 
Indeed, matrix and tensor completion is only meaningful with missing entries. Fortunately, the equivariance properties of \cref{thm:pe_layer} continue to hold when we only consider the nonzero (observed) entries. For sparse matrices, we continue to use the same parameter-sharing scheme across rows and columns, with the only difference being that we limit the model to observed entries. We now make this precise.

Let $\XX \in \Re^{N \times M \times K}$ be a sparse exchangeable array with $K$ channels, where \textit{all} the channels for each row-column pair $\langle n,m \rangle$ are either fully observed or completely missing.
Let $\ind$ identify the set of such non-zero indices.
Let $\row_n = \{ m \mid (n,m) \in \ind \}$ be the non-zero entries in the $n^{th}$ row of $\XX$, and let $\col_m$ be the non-zero entries of its $m^{th}$ column. For this sparse matrix, the terms in the layer of \cref{eq:layer_multi_channel_mat} are adapted as one would expect:
\begin{align*}\label{eq:layer_multi_channel_sparse}%
&\frac{w_2\chn{k,o}}{N} \sum_{n'} \XX_{n',m}\chn{k}  \quad \rightarrow \quad  \frac{w_2\chn{k,o}}{|\col_m|} \sum_{n' \in \col_m} \XX_{n',m}\chn{k}\\
&\frac{w_3\chn{k,o}}{M} \sum_{m'} \XX_{n,m'}\chn{k} \quad \rightarrow \quad 
\frac{w_3\chn{k,o}}{|\row_n|}\sum_{m' \in \row_n} \XX_{n,m'}\chn{k}\\
&\frac{w_4\chn{k,o}}{N M}\sum_{n',m'} \XX_{n',m'}\chn{k}   \quad \rightarrow \quad \frac{w_4\chn{k,o}}{|\ind|}\sum_{(n',m') \in \ind} \XX_{n',m'}\chn{k}
\end{align*}

\section{Matrix Factorization and Completion}\label{sec:fact_comp}
Recommender systems are very widely applied, with many modern applications suggesting new items (\eg movies, friends, restaurants, etc.) to users based on previous ratings of other items. 
The core underlying problem is naturally posed as a matrix completion task: each row corresponds to a user and each column corresponds to an item; the matrix has a value for each rating given to an item by a given user; the goal is to fill in the missing values of this matrix. 

In \cref{sec:ind_trans} we review two types of analysis in dealing with exchangeable matrices.
\cref{sec:architectures} introduces two architectures: a self-supervised model---a simple composition of exchangeable matrix layers---that is trained to produce randomly removed entries of the observed matrix during the training; and a factorized model that uses an auto-encoding nonlinear factorization scheme. While there are innumerable methods for (nonlinear) matrix factorization and completion, both of these models are the first to generalize to inductive settings while achieving competitive performance in transductive settings. 
\cref{sec:subsampling} introduces two subsampling techniques for large sparse matrices followed by a literature review in \cref{sec:related}. 

\subsection{Inductive and Transductive Analysis}\label{sec:ind_trans}
In matrix completion, during training we are given a sparse input matrix $\XX_{\text{tr}}$ with observed entries $\ind_{\text{tr}} = \{(n,m)\}$. At test time, we are given $\XX_{\text{ts}}$ with observed entries $\ind_{\text{ts}} = \{(n',m')\}$, and we are interested in predicting (some of) the missing entries of $\XX_{\text{ts}}$, expressed through $\ind'_{\text{ts}}$.
In the transductive or \textit{matrix interpolation} setting, $\ind_{\text{tr}}$ and $\ind_{\text{ts}}$ have overlapping rows and/or columns---that is, at least one of the following is true:
$\{m \mid (n,m) \in \ind\} \cap \{m' \mid (n',m') \in \ind'\} \neq \emptyset$ or
$\{n \mid (n,m) \in \ind\} \cap \{n' \mid (n',m') \in \ind'\} \neq \emptyset$.
In fact, often $\XX_{\text{tr}}$ and $\XX_{\text{ts}}$ are identical.
Conversely, in the inductive or \textit{matrix extrapolation} setting, we are interested in making predictions about completely unseen entries: the training and test row/column indices are completely disjoint.
We will even consider cases where $\XX_{\text{tr}}$ and $\XX_{\text{ts}}$ are completely different datasets---\eg movie-rating vs music-rating.
The same distinction applies in matrix factorization. Training a model to factorize a 
particular given matrix is transductive, while factorizing unseen 
matrices \emph{after} training is inductive. 

\subsection{Architectures}\label{sec:architectures}
\begin{figure}[t]\centering
\includegraphics[width=\columnwidth,trim=0px 72px 0px 70px,clip=true]{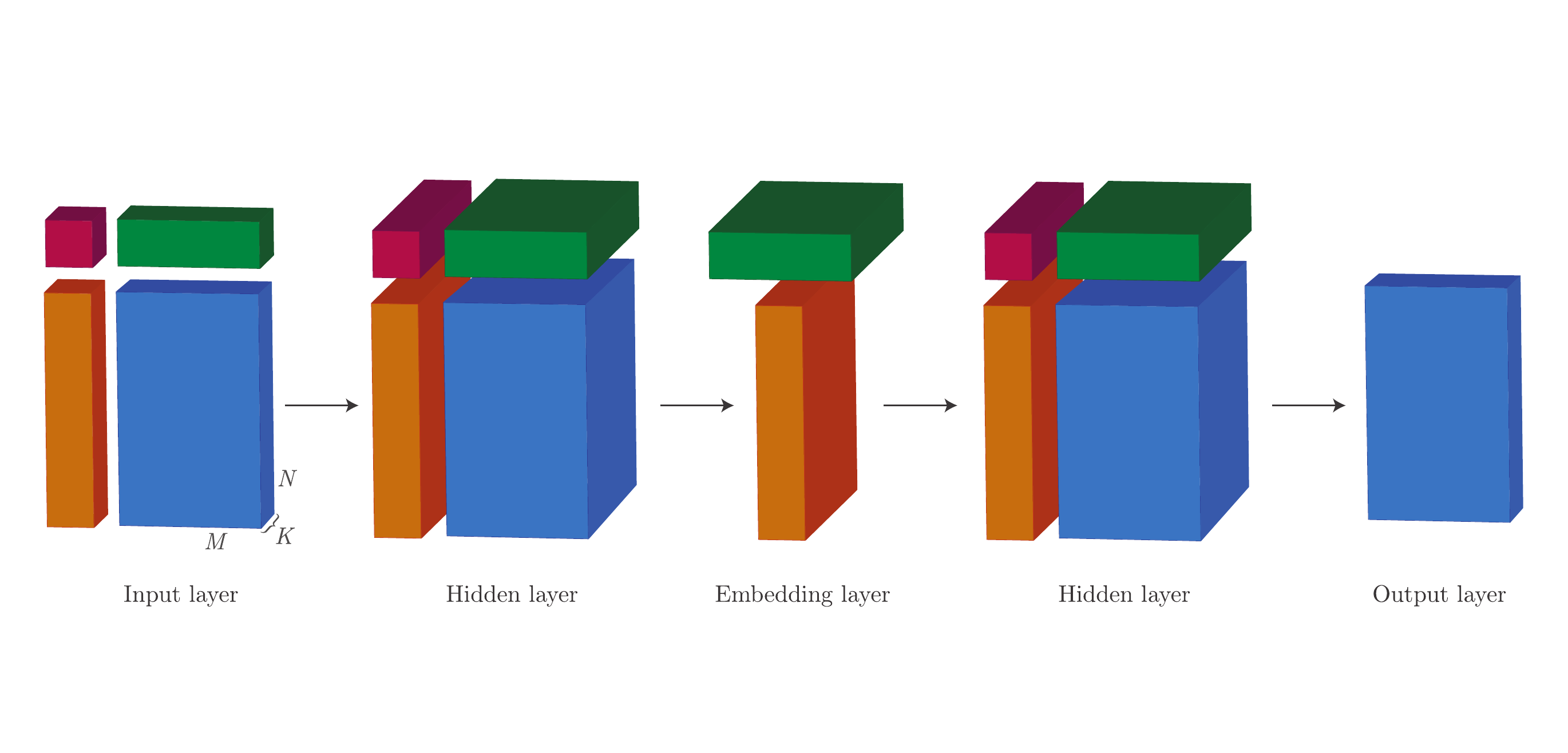}
\vspace{-1.3em}\caption{Factorized exchangeable autoencoder. The encoder maps from the input tensor to an embedding layer of row / column factors via one or more hidden layers. The decoder attempts to reconstruct the input using the factors via one or more hidden layers.}
\label{fig:fact_ae}
\end{figure}

\paragraph{Self-supervised Exchangeable Model} When the task is matrix completion, a simple deep yet PE model is a 
composition of exchangeable matrix layers.
That is the function $f_{ss}: \Re^{N \times M \times K} \to \Re^{N \times M \times K}$ is simply a composition of exchangeable matrix layers. Given the matrix $\XX$ with observed entries $\ind$,
we divide $\ind = \ind_{\text{in}} \cup \ind_{\text{pr}}$ into disjoint input and prediction entries. 
We then train $f_{ss}(\XX_{\text{in}})$ to predict the prediction entries $\XX_{\text{pr}}$.

\vspace{-.6em minus .3em}\paragraph{Factorized Exchangeable Autoencoder (FEA)}
Our factorized autoencoder is composed of an encoder and a decoder. 
The encoder $f_{\text{enc}}:\Re^{N\times M \times K} \to \Re^{K_N \times N} \times \Re^{K_M \times M}$ maps
the (sparse) input matrix $\XX \in \Re^{N \times M \times K}$ to a row-factor $\ZZ_{N} \in \Re^{K_N \times N}$ and a column-factor $\ZZ_{M} \in \Re^{K_M \times M}$. To do so, the encoder uses a composition of
exchangeable matrix layers. The output of the final layer $\YY^{l} \in \Re^{N \times M \times K^{l}}$ is pooled across rows and columns (and optionally passed through a feed-forward layer) to 
produce latent factors $\ZZ_{N}$ and $\ZZ_{M}$.
The decoder $g_{\text{dec}}:\Re^{K_N \times N} \times \Re^{K_M \times M} \to \Re^{N\times M \times K}$ also uses a composition of exchangeable matrix layers, and reconstructs the input matrix $\XX$ from the factors. The optimization objective is to minimize reconstruction error
$\ell(\XX, g_{\text{dec}}(f_{enc}(\XX)))$; similar to classical auto-encoders. 

This procedure is also analogous to classical matrix factorization, with an an important distinction that
once trained, we can readily factorize the unseen matrices without performing any optimization. Note that the same architecture trivially extends to tensor-factorization, where we use an exchangeable tensor layer (see \cref{sec:tensors}). 

\vspace{-.6em minus .3em}\paragraph{Channel Dropout}
Both the factorized autoencoder and self-supervised exchangeable model are flexible enough to make regularization important for good generalization performance. Dropout \citep{srivastava2014dropout} can be extended to apply to exchangeable matrix layers by noticing that each channel in an exchangeable matrix layer is analogous to a single unit in a standard feed-forward network. We therefore randomly drop out whole channels during training (as opposed to dropping out individual elements). This procedure is equivalent to the \emph{SpatialDropout} technique used in convolutional networks  \citep{tompson2015efficient}.

\subsection{Subsampling in Large Matrices}\label{sec:subsampling}
A key practical challenge arising from our approach is that our models are designed to take the whole data matrix $\XX$ as input and will make different (and typically worse) predictions if given only a subset of the data matrix. As datasets grow, the model and input data become too large to fit within fixed-size GPU memory. This is problematic both during training and at evaluation time because our models rely on aggregating shared representations across data points to make their predictions. To address this, we explore two subsampling procedures. 
\vspace{-.6em minus .3em}\paragraph{Uniform sampling} The simplest approach is to sample sub-matrices of $\XX$ by uniformly sampling from its (typically sparse) elements. 
This has the advantage that each submatrix is an unbiased sample of the full matrix and that the procedure is computationally cheap, but has the potential to limit the performance of the model because the relationships between the elements of $\XX$ are sparsified.

\begin{figure}[h]\centering
\includegraphics[width=\columnwidth]{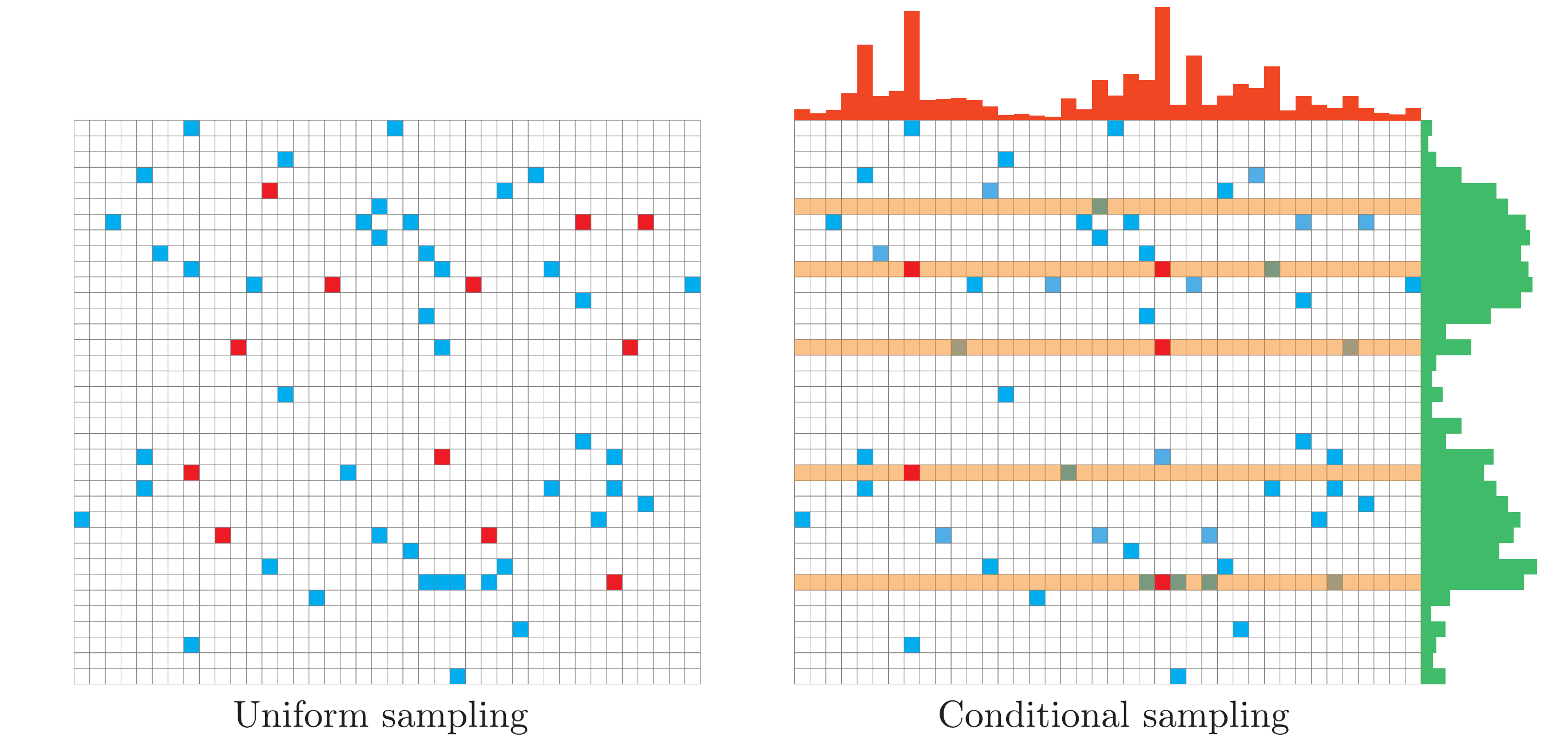}
\caption{Uniform sampling (left) selects samples (red) uniformly from the non-zero indices of the the matrix $\XX$ while conditional sampling (right) first samples a set of rows (shown in orange) from the row marginal distribution (green) and then selects sample from the resulting column conditional distribution.}
\label{fig:sampling}
\end{figure}

\vspace{-.6em minus .3em}\paragraph{Conditional sampling} Rather than sparsifying interactions between all set members, we can pick a subset of rows and columns and maintain all their interactions; see Figure \ref{fig:sampling}. This procedure is unbiased as long as each element $(n,m) \in \ind$ has the same probability of being sampled.
To achieve this, we first sample a subset of rows $\set{N}' \subseteq \set{N}=\{1,\ldots,N\}$ from the marginal $P(n) := \frac{|\row_{n}|}{|\ind|}$, followed by subsampling of colums using the marginal distribution over the columns, within the selected rows: $P(m \mid \set{N}') = \frac{|\{(m,n) \in \ind \mid n \in \set{N}'\}|}{\{(m',n) \in \ind \mid n \in \set{N}'\}|}$. This gives us a set of columns $\set{M}' \subseteq \set{M}$. We consider any observation within $\set{N}' \times \set{M}'$ as our subsample: $\ind_{\text{sample}} := \{(n,m) \in \ind \mid n \in \set{N}', m \in \set{M}'\}$. 
This sampling procedure is more expensive than uniform sampling, as we have to calculate 
conditional marginal distributions for each set of samples.

\paragraph{Sampling at test time} At training time all we must ensure is that we sample in an unbiased way; however, at test time we need to ensure that the test indices $\ind_{\text{ts}}$ of large test matrix $\XX_{\text{ts}}$ are all sampled.
Fortunately, according to the \textit{coupon collectors' lemma}, in expectation we only need to repeat
random sampling of indices 
$\log(|\ind_{\text{ts}}|)$ times more ($\approx 10 \times$ in practice) than an ideal partitioning of $\ind_{\text{ts}}$, in order to cover all the relevant indices.

\subsection{Related Literature}\label{sec:related}
The literature in matrix factorization and completion is vast. 
The classical approach to solving the matrix completion problem is to find some low rank (or sparse) approximation that minimizes a reconstruction loss for the observed ratings \citep[see e.g.,][]{candes2009exact, koren2009matrix, mnih2008probabilistic}. 
Because these procedures learn embeddings for each user and item to make predictions, they are {transductive}, meaning they can only make predictions about users and items observed during training. To our knowledge this is also true for all recent   
deep factorization and other collaborative filtering techniques~\citep[\eg][]{salakhutdinov2007restricted,deng2017factorized,sedhain2015autorec,wang2015collaborative,li2015deep,zheng2016neural,dziugaite2015neural}.
An exception is a recent work by \citet{yang2016revisiting} that extends factorization-style approaches to the {inductive} setting (where predictions can be made on unseen users and items). However their method relies on additional side information to represent users and items. By contrast, our approach is able to make inductive completions on rating matrices that may differ from that which was observed during training without using any side information (though our approach can easily incorporate side information).

Matrix completion may also be posed as predicting edge weights in bipartite graphs~\citep{berg2017graph,monti_geomatrix}.
This approach builds on recent work applying convolutional neural networks to graph-structured data \citep{scarselli2009graph,bruna2013spectral,duvenaud2015convolutional, defferrard2016convolutional, kipf2016semi,hamilton2017inductive}. 
As we saw, parameter sharing in graph convolution is closely related to parameter sharing in exchangeable matrices, and indeed it is a special case where $w_2 = w_3$ in Equation~\cref{eq:w}.

\section{Empirical Results}\label{sec:empirical}
For reproducibility we have released Tensorflow and Pytorch implementations of our model.\footnote{\noindent Tensorflow: \url{https://github.com/mravanba/deep_exchangeable_tensors}. Pytorch: \url{https://github.com/jhartford/AutoEncSets}.} 
Details on the training and architectures appear in the appendix. 
\cref{sec:results_trans} reports experimental results in the standard transductive (matrix interpolation)
setting. However, more interesting results are reported in \cref{sec:results_inductive}, where 
we test a trained deep model on a completely different dataset. Finally \cref{sec:results_sampling} compares the
model's performance when using different sampling procedures. The datasets used in our experiments are summarized in \cref{results:datasets}.

\begin{table}[h]
\tablefontsmaller
\begin{tabular}{l r r r r} 
\toprule
\textbf{Dataset} & \textbf{Users} & \textbf{Items} & \textbf{Ratings} & \textbf{Sparsity} \\ [0.5ex] 
\midrule
MovieLens 100K & 943 & 1682 & 100,000 & 6.30\% \\ 
MovieLens 1M & 6040 & 3706 & 1,000,209 & 4.47\%\\
Flixster & 3000& 3000& 26173  & 0.291\%\\
Douban & 3000& 3000& 136891 & 1.521\%\\
Yahoo Music& 3000& 3000 & 5335 & 0.059\%\\
Netflix & 480,189 & 17,770 & 100,480,507 & 1.178\%\\
\bottomrule
\end{tabular}
\caption {Number of users, items and ratings for the  data sets used in our experiments. MovieLens data sets are standard \citep{movielens}, as is Netflix, while for Flixster, Douban and Yahoo Music we used the $3000 \times 3000 $ submatrix presented by \citep{monti_geomatrix} for comparison purposes.}\label{results:datasets}
\end{table}

\subsection{Transductive Setting (Matrix Interpolation)}\label{sec:results_trans}
Here, we demonstrate that exploiting the PE structure of the exchangeable matrices
allows us to achieve results competitive with state-of-the-art, 
while maintaining a constant number of parameters.
Note that the number of parameters in all the competing methods grow with $N$ and/or $M$.


In \cref{results:perf100k} we report that the self-supervised exchangeable model is able to achieve state of the art performance on 
MovieLens-100K dataset. For MovieLens-1M dataset, we cannot fit the whole dataset into the GPU memory for training and therefore use conditional subsampling; also see \cref{sec:results_sampling}. Our results on this dataset are summarized in table \cref{results:perf1M}. On this larger dataset both models gave comparatively weaker performance than what we observed on the smaller ML-100k dataset and in the extrapolation results. We suspect this is largely due to memory constraints: there is a trade-off between the size of the model (in terms of number of layers and units per layer) and the batch size one can train. We found that both larger batches and deeper models tended to offer better performance, but on these larger datasets it is not currently possible to have both.
The results for the factorized exchangeable autoencoder architecture are similar and reported in the same table.

\begin{table}[h]
\tablefont
\begin{tabular}{l r} 
  \toprule
  \textbf{Model} & \textbf{ML-100K} \\ [0.5ex] 
  \midrule
  MC {\smaller{\citep{candes2009exact}}} & 0.973 \\
  GMC {\smaller{\citep{kalofolias2014completion}}}  & 0.996 \\
  GRALS {\smaller{\citep{rao2015collab}}}  & 0.945 \\
  sRGCNN {\smaller{\citep{monti_geomatrix}}} & 0.929 \\
  Factorized EAE (ours)&0.920 \\
  GC-MC {\smaller{\citep{berg2017graph}}} & \textbf{0.910} \\
  Self-Supervised Model (ours) & 0.910 \\
\bottomrule
\end{tabular}
\caption {Comparison of RMSE scores for the MovieLens-100k dataset, based on the canonical 80/20 training/test split. Baseline numbers are taken from \citep{berg2017graph}.} \label{results:perf100k}
\end{table}

\begin{table}[h]
\tablefont
\begin{tabular}{l r} 
  \toprule
  \textbf{Model} & \textbf{ML-1M}  \\ [0.5ex] 
  \midrule
  PMF {\smaller{\citep{mnih2008probabilistic}}} & 0.883  \\
  Self-Supervised Model (ours)& 0.863 \\
  Factorized EAE (ours)& 0.860\\
  I-RBM {\smaller{\citep{salakhutdinov2007restricted}}}  & 0.854 \\
  BiasMF {\smaller{\citep{koren2009matrix}}}  & 0.845 \\
  NNMF {\smaller{\citep{dziugaite2015neural}}} & 0.843 \\
  LLORMA-Local {\smaller{\citep{lee2013local}}} & 0.833 \\
  GC-MC {\smaller{\citep{berg2017graph}}} & 0.832\\
  I-AUTOREC {\smaller{\citep{sedhain2015autorec}}} & 0.831 \\
  CF-NADE {\smaller{\citep{zheng2016neural}}} & \textbf{0.829} \\
  \bottomrule
\end{tabular}
\caption {Comparison of RMSE scores for the MovieLens-1M dataset on random 90/10 training/test split. Baseline numbers are taken from \citep{berg2017graph}.}\label{results:perf1M}
\end{table}

\subsection{Inductive Setting (Matrix Extrapolation)}\label{sec:results_inductive}
Because our model does not rely on any per-user or per-movie parameters, it should be able to generalize to new users and movies that were not present during training. We tested this by training an exchangeable factorized autoencoder on the MovieLens-100k dataset and then evaluating it on a subsample of data from the MovieLens-1M dataset. At test time, the model was shown a portion of the new ratings and then made to make predictions on the remaining ratings. 

\begin{figure}[h]\centering
\includegraphics[width=0.95\columnwidth, trim=0px 0px 0px 0px,clip=true]{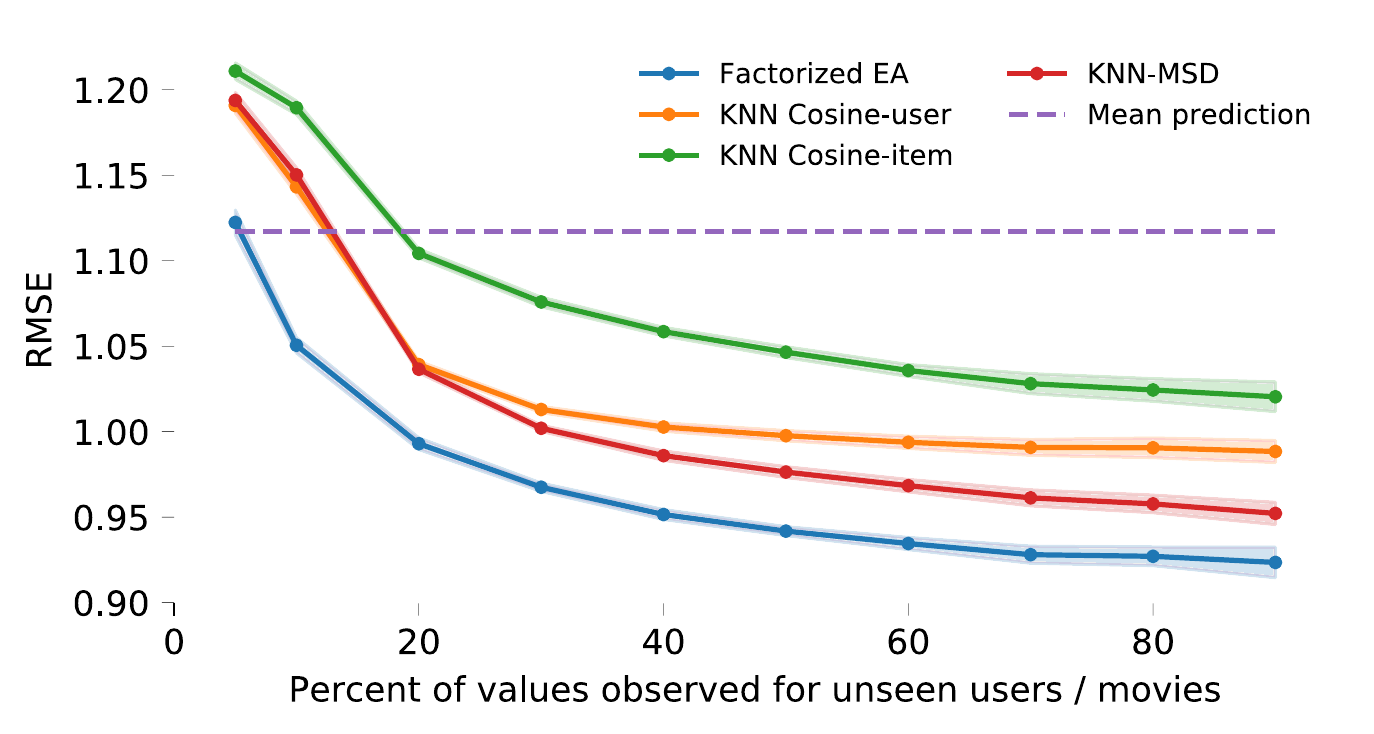}
\caption{Evaluation of our model's ability to generalize. We trained on ML-100k and evaluated on a subset of the ML-1M data. At evaluation time, $p\%$ of the ML-1M data was treated as \emph{observed} and the model was required to complete the remaining $(1-p)\%$ ($p$ varied from $5\%$ to $95\%$). The model outperforms nearest-neighbour approaches for all values of $p$ and degrades gracefully to the baseline of predicting the mean in the small data case.}
\label{fig:missing}
\end{figure}

Figure \ref{fig:missing} summarizes the results where we vary the amount of data shown to the model from 5\% of the new ratings up to 95\% and compare against K-nearest neighbours approaches. Our model significantly outperforms the baselines in this task and performance degrades gracefully as we reduce the amount of data observed.

\vspace{-.6em minus .3em}\paragraph{Extrapolation to new datasets}
Perhaps most surprisingly, we were able to achieve competitive results when training and testing on completely disjoint datasets. For this experiment we stress-tested our model's inductive ability by testing how it generalizes to new datasets \emph{without retraining}. We used a Factorized Exchangeable Autoencoder that was trained to make predictions on the MovieLens-100k dataset and tasked it with making predictions on the Flixster, Douban and YahooMusic datasets. We then evaluated its performance against models trained for each of these individual datasets. All the datasets involve rating prediction tasks, so they share some similar semantics with MovieLens, but they have different user bases and (in the case of YahooMusic) deal with music not movie ratings, so we may expect some change in the rating distributions and user-item interactions. Furthermore, the Flixster ratings are in 0.5 point increments from $1-5$ and YahooMusic allows ratings from $1 - 100$, while Douban and MovieLens ratings are on $1-5$ scale. To account for the different rating scales, we simply binned the inputs to our model to a $1 - 5$ range and, where applicable, linearly rescaled the output before comparing it to the true rating\footnote{Because of this binning procedure, our model received input data that is considerably coarser-grained than that which was used for the comparison models.}. Despite all of this, Table \ref{results:generalize} shows that our model achieves very competitive results with models that were trained for the task. 

For comparison, we also include the performance of a Factorized EAE trained on the respective datasets. This improves performance of our model over previous state of the art results on the Flixster and YahooMusic datasets and gives very similar performance to \citet{berg2017graph}'s GC-MC model on the Douban dataset. Interestingly, we see the largest performance gains over existing approaches on the datasets in which ratings are sparse (see Table \ref{results:datasets}).


\subsection{Comparison of sampling procedures}
\label{sec:results_sampling}

\begin{figure}[h]\centering
\includegraphics[width=0.95\columnwidth, trim=0px 0px 0px 0px,clip=true]{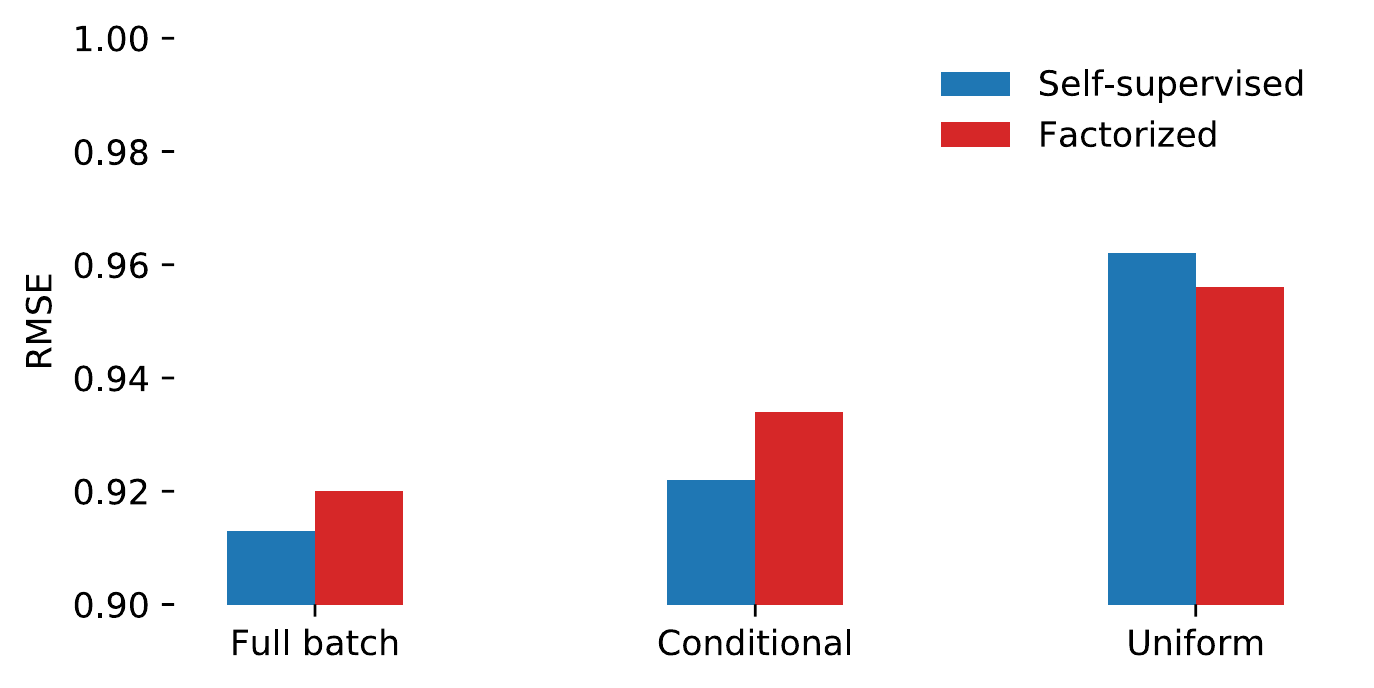}
\caption{Performance difference between sampling methods on the ML-100k. The two sampling methods use minibatches of size 20 000, while the full batch method used all 75 000 training examples. Note that the y-axis does not begin at 0.}
\label{fig:sampling_perf}
\end{figure}

We evaluated the effect of subsampling the input matrix $\XX$ on performance, for the MovieLens-100k dataset. The results are summarized in Figure \ref{fig:sampling_perf}. 
The two key findings are: I) even with large batch sizes of 20 000 examples, performance for both sampling methods is diminished compared to the full batch case. We suspect that our models' weaker results on the larger ML-1M dataset may be partially attributed to the need to subsample. II) the conditional sampling method was able to recover some of the diminished performance. We believe it is likely that more sophisticated sampling schemes that explicitly take into account the dependence between hidden nodes will lead to better performance but we leave that to future work.



\begin{table}[t]
  \tablefontsmaller
  \smaller{\begin{tabular}{l r r r r} 
  \toprule
  \textbf{Model} & \rotatebox{90}{\textbf{Flixster}} & \rotatebox{90}{\textbf{Douban}} & \rotatebox{90}{\textbf{YahooMusic}} & \rotatebox{90}{\textbf{Netflix}} \\ [0.5ex] 
  \midrule
  GRALS {\smaller{\citep{rao2015collab}}}  &1.313 & 0.833 & 38.0 & - \\
  sRGCNN {\smaller{\citep{monti_geomatrix}}} & 1.179 &  0.801 & 22.4 & - \\
  GC-MC {\smaller{\citep{berg2017graph}}} & 0.941 & \textbf{0.734} & 20.5 & - \\
  Factorize EAE (ours) & \textbf{0.908} & 0.738 & \textbf{20.0} & -\\
Factorize EAE {\smaller{\color{red}{(trained on ML100k)}}} & 0.987 & 0.766 & 23.3 & 0.918\\
  Netflix Baseline & - & - & - & 0.951 \\
  PMF {\smaller{\citep{mnih2008probabilistic}}} & - & - & - & 0.897 \\
  LLORMA-Local {\smaller{\citep{lee2013local}}} & - & - & - & 0.834 \\
  I-AUTOREC {\smaller{\citep{sedhain2015autorec}}} & - & - & - & 0.823 \\
  CF-NADE {\smaller{\citep{zheng2016neural}}} & - & - & - & \textbf{0.803} \\       
  \bottomrule
\end{tabular}}
\caption {Evaluation of our model's ability to generalize across datasets. We trained a factorized model on ML100k and then evaluated it on four new datasets. Results for the smaller datasets are taken from \citep{berg2017graph}. Netflix Baseline shows state of the art prior to the Netflix Challenge.}\label{results:generalize}
\end{table}

\section{Extention to Tensors}\label{sec:tensors}
In this section we generalize the exchangeable matrix layer to higher-dimensional arrays (tensors) and formalize its optimal qualities.
Suppose $\XX \in \Re^{N_1 \times ... \times N_D}$ is our $D$-dimensional data tensor, and $\vec{\XX}$ its vectorized form. We index $\vec{\XX}$ by tuples $(n_1, n_2, ..., n_D)$, corresponding to the original dimensions of $\XX$. The precise method of vectorization is irrelevant, provided it is used consistently. Let $\pN = \prod_i N_i$ and let $\bn = (n_i, n_{-i})$ be an element of ${N_1 \times ... \times N_D}$ such that $n_i$ is the value of the $i$-th entry of $\bn$, and $n_{-i}$ the values of the remaining entries (where it is understood that the ordering of the elements of $\bn$ is unchanged). We seek a layer that is equivariant only to certain, meaningful, permutations of $\vec{\XX}$. This motivates our definition of an exchangeable tensor layer in a manner that is completely analogous to Definition \ref{def:2pe_layer} for matrices. 

For any positive integer $N$, let $\grn{S}{N}$ denote the symmetric group of all permutations of $N$ objects. Then $\grn{S}{N_1} \times ... \times \grn{S}{N_D}$ refers to the product group of all permutations of $N_1$ through $N_D$ objects, while $\grn{S}{\pN}$ refers to the group of all permutations of $\pN=\prod_i N_i$ objects. So $\grn{S}{N_1} \times ... \times \grn{S}{N_D}$ is a proper subgroup of $\grn{S}{\pN}$ having $\prod_i (N_i!)$ members, compared to $(\prod_i N_i)!$ members in $\grn{S}{\pN}$. We want a layer that is equivariant to \textit{only} those permutations in $\grn{S}{N_1} \times ... \times \grn{S}{N_D}$, but not to any other member of $\grn{S}{\pN}$.

\begin{figure}[t]\centering
\includegraphics[width=\columnwidth]{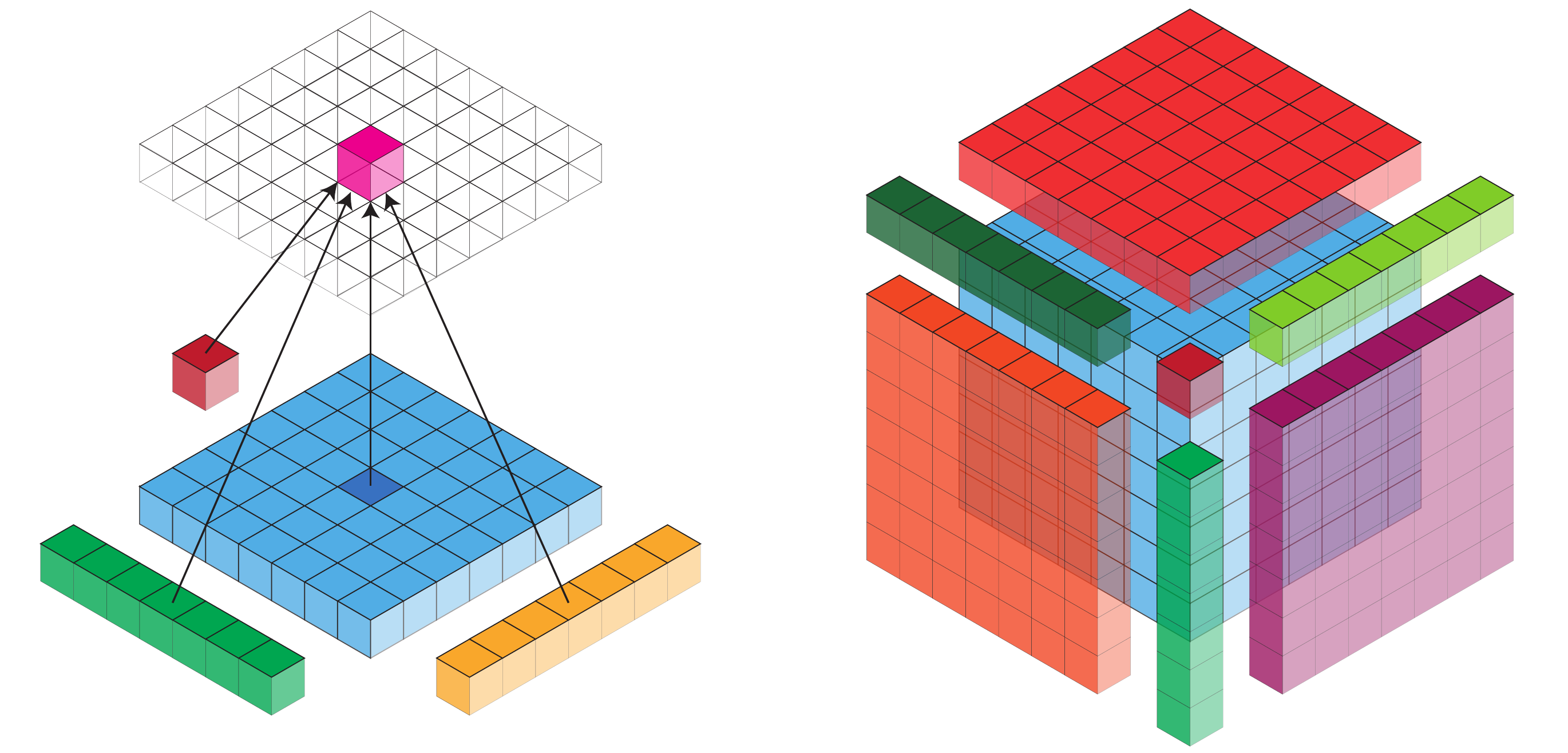}
\caption{Pooling structure implied by the tied weights for matrices (left) and 3D tensors (right). The pink cube highlights one element of the output. It is calculated as a function of the corresponding element from the input (dark blue), pooled aggregations over the rows and columns of the input (green and yellow), and pooled aggregation over the whole input matrix (red). In the tensor case (right), we pool over all sub-tensors (orange and purple sub-matrices, green sub-vectors and red scalar). For clarity, the output connections are not shown in the tensor case.
}
\label{fig:exch_matrix_layer}
\end{figure}

\begin{definition}[exchangeable tensor layer]
Let $\grn{g}{\pN} \in \grn{S}{N_1} \times \grn{S}{N_2} \times ... \times \grn{S}{N_D}$
and $\mat{G}\prm{\pN}$ be the corresponding permutation matrix. 
Then the neural layer $\unvec{\sigma(\WW \vec{\XX})}$ with $\XX \in \Re^{N_{1}\times \ldots \times N_D}$ 
and $\WW \in \Re^{\pN \times \pN}$ is an exchangeable tensor layer if permuting the elements of the input along any set of axes results in the same permutation of the output tensor:
\begin{align}
G\prm{\pN} \sigma(W \vec{X}) = \sigma(W G\prm{\pN} \vec{X}) \quad \forall \XX ,
\end{align}
and moreover for any other permutation of the elements $\XX$ (\ie permutations that are not along axes), 
there exists an input $\XX$ for which this equality does not hold. 
\end{definition}

The following theorem asserts that a simple parameter tying scheme achieves the desired permutation equivariance for tensor-structured data. 

\begin{theorem}\label{thm:equivar}
The layer ${\YY} = \unvec{\sigma( \WW \vec{\XX} )}$, where $\sigma$ is any strictly monotonic, element-wise nonlinearity (\eg sigmoid), is an exchangeable tensor layer iff the parameter matrix $\WW \in \Re^{\pN \times \pN}$ is defined as follows. 

For each $\set{S} \subseteq [D] = \{1,\ldots,D\}$, define a distinct parameter $w_\set{S} \in \Re$, and tie the 
entries of $\WW$ as follows 
\begin{align} \label{eqn:params_tensor}
  \WW_{\bn, \bn'} &:= w_\set{S} \quad \text{s.t.} \quad n_i = n_i' \iff i \in \set{S} .
\end{align}
That is, the $(\bn,\bn')$-th element of $\WW$ is uniquely determined by the set of indices at which $n$ and $n'$ are equal. 
\end{theorem}

In the special case that $\XX \in \Re^{N_1 \times N_2}$ is a matrix, this gives the formulation of $\WW$ described in Section \ref{sec:layer}. 
Theorem \ref{thm:equivar} says that a layer constructed in this manner is equivariant with respect to \textit{only} those permutations of $\pN$ objects that correspond to permutations of  sub-tensors along the $D$ dimensions of the input. The proof is in the Appendix. 
Equivariance with respect to permutations in $\grn{S}{N_1} \times \grn{S}{N_2} \times ... \times \grn{S}{N_D}$) follows as a simple corollary of Theorem 2.1 in \cite{ravanbakhsh_symmetry}. Non-equivariance with respect to other permutations follows from the following Propositions. 
\begin{proposition}\label{prop:equivar1}
Let $\grn{g}{\pN} \in \grn{S}{\pN}$ be an ``illegal'' permutation of elements of the tensor $X$ -- that is $\grn{g}{\pN} \not\in \grn{S}{N_1} \times \grn{S}{N_2} \times ... \times \grn{S}{N_D}$. Then there exists a dimension $i \in [D]$ such that, for some $n_i, n_i', n_{-i}, n_{-i}'$:
    \begin{align*}
      \grn{g}{\pN}((n_i, n_{-i})) &= (n_i', n_{-i}), \quad \text{but} \\
      \grn{g}{\pN}((n_i, n_{-i}')) &\neq (n_i', n_{-i}'). 
    \end{align*}
\end{proposition}

If we consider the sub-tensor of $\XX$ obtained by fixing the value of the $i$-th dimension to $n_i$, we expect a ``legal'' permutation to move this whole subtensor to some $n_i'$ (it could additionally shuffle the elements within this subtensor.) This Proposition asserts that an ``illegal'' permutation $\grn{g}{\pN}\not\in \grn{S}{N_1} \times \grn{S}{N_2} \times ... \times \grn{S}{N_D}$ is guaranteed to violate this constraint for some dimension/subtensor combination. 
The next proposition asserts that if we can identify such inconsistency in permutation of sub-tensors then we can
identify and entry in $ \mat{G}\prm{\pN} \WW $ that will differ from  $\WW \mat{G}\prm{\pN}$, and therefore
for some input tensor $X$, the equivariance property is violated -- \ie $G\prm{\pN} \sigma(W \vec{X}) \neq \sigma(W G\prm{\pN} \vec{X})$.

\begin{proposition}\label{prop:equivar2}
  Let $\grn{g}{\pN} \in \grn{S}{\pN}$ with $\mat{G}\prm{\pN} \in \{0,1\}^{\pN \times \pN}$ the corresponding permutation matrix. Suppose $\grn{g}{\pN} \not\in \grn{S}{N_1} \times \grn{S}{N_2} \times ... \times \grn{S}{N_D}$, and let $\WW \in \Re^{\pN \times \pN}$ be as defined above. If there exists an $i \in [D]$, and some $n_i, n_i', n_{-i}, n_{-i}'$ such that 
    \begin{align*}
      \grn{g}{\pN}((n_i, n_{-i})) &= (n_i', n_{-i}), \quad \text{and} \\
      \grn{g}{\pN}((n_i, n_{-i}')) &\neq (n_i', n_{-i}'), 
    \end{align*}
    then
    \begin{align*}
      \big( \mat{G}\prm{\pN} \WW \big)_{(n_i',n_{-i}'), (n_i, n_{-i})} \neq \big( \WW \mat{G}\prm{\pN} \big)_{(n_i', n_{-i}'), (n_i, n_{-i})}
    \end{align*}
\end{proposition}

Proposition \ref{prop:equivar2} makes explicit a particular element at which the products $\mat{G}\prm{N} \WW$ and $\WW \mat{G}\prm{N}$ will differ, provided $\grn{g}{\pN}$ is not of the desired form.

Theorem \ref{thm:equivar} shows that this parameter sharing scheme produces a layer that is equvariant to exactly those permutations we desire, and moreover, it is optimal in the sense that any layer having fewer ties in its parameters (\ie more parameters) would fail to be equivariant.  

\section*{Conclusion}\label{sec:discussion}
This paper has considered the problem of predicting relationships between the elements of two or more distinct sets of objects, where the data can also be expressed as an exchangeable matrix or tensor. We introduced a weight tying scheme enabling the application of deep models to this type of data. We proved that our scheme always produces permutation equivariant models and that no increase in model expressiveness is possible without violating this guarantee. Experimentally, we showed that our models achieve state-of-the-art or competitive performance on widely studied matrix completion benchmarks. Notably, our models achieved this performance despite having a number of parameters independent of the size of the matrix to be completed, unlike \emph{all} other approaches that offer strong performance. Also unlike these other approaches, our models can achieve competitive results for the problem of \emph{matrix extrapolation}: asking an already-trained model to complete a new matrix involving new objects that were unobserved at training time. 
Finally, we observed that our methods were surprisingly strong on various transfer learning tasks: extrapolating from MovieLens ratings to Fixter, Douban, and YahooMusic ratings. All of these contained different user populations and different distributions over the objects being rated; indeed, in the YahooMusic dataset, the underlying objects were not even of the same kind as those rated in training data.

\section*{Acknowledgment}\label{sec:ack}
We want to thank the anonymous reviewers for their constructive feedback.
This research was enabled in part by support provided by NSERC Discovery Grant, WestGrid and Compute Canada.

\bibliography{refs.bib}
\bibliographystyle{icml2018}

\newpage
\clearpage
\appendix

\section{Notation}
\begin{itemize}
\item $\XX \in \Re^{N_1 \times ... \times N_D}$, the data tensor
\item $\xx \in \Re^{\prod_i N_i}$, vectorized $\XX$, also denoted by $\vec{\XX}$.
\item $[N]$: the sequence $\{n\}_{n=1,...,N} = (1, 2, ... N)$
\item $[N_1 \times ... \times N_D]$, the sequence $\{(n_1, ..., n_D)\}_{n_1 \in [N_1], ..., n_D \in [N_D]}$ of $D$-dimensional tuples over $N_1 \times ... \times N_D$
\item $\bn = (n_i, n_{-i})$, an element of ${N_1 \times ... \times N_D}$
\item $\pN=\prod_i N_i$
\item $\grn{S}{N}$, symmetric group of \textit{all} permutations of $N$ objects
\item $\grn{G}{N} = \{\grn{g}{N}_i\}_i$, a permutation group of $N$ objects 
\item $\grn{g}{N}_i$ or $\mat{G}_i\prm{N}$, both can refer to the matrix form of the permutation $\grn{g}{N}_i \in \grn{G}{N}$.
\item $\grn{g}{N}_i(n)$ refers to the result of applying $\grn{g}{N}_i$ to $n$, for any $n \in [N]$.
\item $\sigma$, an arbitrary, element-wise, strictly monotonic, nonlinear function such as sigmoid.
\end{itemize}

\section{Proofs}\label{app:proofs}

\subsection{Proof of Proposition \ref{prop:equivar1}}
\begin{proof}
Let $\XX \in \Re^{N_1 \times ... \times N_D}$ be the data matrix. We prove the contrapositive by induction on $D$, the dimension of $\XX$. Suppose that, for all $i \in [D]$ and all $n_i, n_i', n_{-i}, n_{-i}'$, we have that $\grn{g}{\pN}((n_i, n_{-i})) = (n_i', n_{-i})$ implies $\grn{g}{\pN}((n_i, n_{-i}')) = (n_i', n_{-i}')$. This means that, for any $\bn = (n_i, n_{-i}) = (n_1, ..., n_i, ..., n_D)$ the action $\grn{g}{\pN}$ takes on $n_i$ is independent of the action $\grn{g}{\pN}$ takes on $n_{-i}$, the remaining dimensions. Thus we can write 
  \begin{align*}
    \grn{g}{\pN}(\bn) = \grn{g}{N_i}(n_i) \times \grn{g}{\pN/N_i}(n_{-i})
  \end{align*}
  Where it is understood that the order of the group product is maintained (this is a slight abuse of notation). If $D=2$ (base case) we have $\grn{g}{\pN}((n_1, n_2)) = \grn{g}{N_1}(n_1) \times \grn{g}{N_2}(n_2)$. So $\grn{g}{\pN} \in \grn{S}{N_1} \times \grn{S}{N_2}$, and we are done. Otherwise, an inductive argument on $\grn{g}{\pN/N_i}$ allows us to write $\grn{g}{\pN}(\bn) = \grn{g}{N_1}(n_1) \times \grn{g}{N_2}(n_2) \times ... \times \grn{g}{N_D}(n_D)$. And so $\grn{g}{\pN} \in \grn{S}{N_1} \times \grn{S}{N_2} \times ... \times \grn{S}{N_D}$, completing the proof.
\end{proof}

\subsection{Proof of Proposition \ref{prop:equivar2}}
\begin{proof}
First, observe that 
\begin{align*}
  \grn{g}{\pN}(\bn) = \bn' &\iff \big( \mat{G}\prm{\pN}  \big)_{\bn', \bn} = 1
\end{align*}
  Now, let $i \in [D]$ be such that, for some $n_i, n_i', n_{-i}, n_{-i}'$ we have
    \begin{align*}
      \grn{g}{\pN}((n_i, n_{-i})) &= (n_i', n_{-i}), \quad \text{and} \\
      \grn{g}{\pN}((n_i, n_{-i}')) &\neq (n_i', n_{-i}'). 
    \end{align*}
  Then by the observation above we have:
    \begin{align*}
      \big( \mat{G}\prm{\pN}  \big)_{(n_i', n_{-i}), (n_i, n_{-i})} &= 1, \quad \text{and} \\
      \big( \mat{G}\prm{\pN}  \big)_{(n_i', n_{-i}'), (n_i, n_{-i}')} &\neq 1.
    \end{align*}
  And so: 
  \begin{align*}
    \big( \mat{G}\prm{\pN} \WW &\big)_{(n_i',n_{-i}'), (n_i, n_{-i})} = \big( \mat{G}\prm{\pN} \big)_{(n_i',n_{-i}'), *} \big(\WW \big)_{*, (n_i, n_{-i})} \\
    &= \sum_{\bi \in [N_1 \times ... \times N_D]} \big( \mat{G}\prm{\pN} \big)_{(n_i',n_{-i}'), \bi} \big(\WW \big)_{\bi, (n_i, n_{-i})} \\
    & \neq \WW_{(n_i, n_{-i}'), (n_i, n_{-i})} 
  \end{align*}
  The last line follows from the observation above and the fact that $\mat{G}\prm{\pN}$ is a permutation matrix and so has only one 1 per row. Similarly, 
  \begin{align*}
    \big( \WW \mat{G}\prm{\pN} &\big)_{(n_i', n_{-i}'), (n_i, n_{-i})} = \big( \WW \big)_{(n_i', n_{-i}'), *} \big( \mat{G}\prm{\pN} \big)_{*, (n_i, n_{-i})} \\
    &= \sum_{\bi \in [N_1 \times ... \times N_D]} \big( \WW \big)_{(n_i', n_{-i}'), \bi} \big( \mat{G}\prm{\pN} \big)_{\bi, (n_i, n_{-i})} \\
    &= \big( \WW \big)_{(n_i', n_{-i}'), (n_i', n_{-i})} 
  \end{align*} 
  Where again the last line follows from the above observation. Now, consider $\WW_{(n_i, n_{-i}'), (n_i, n_{-i})} $ and $\WW_{(n_i', n_{-i}'), (n_i', n_{-i})}$. Observe that $(n_i, n_{-i}')$ differs from $(n_i, n_{-i})$ at exactly the same indices that $(n_i', n_{-i}')$ differs from $(n_i', n_{-i})$. Let $\set{S} \subseteq [D]$ be the set of indices at which $n_{-i}$ differs from $n_{-i}'$. We therefore have 
  \begin{align*}
    \WW_{(n_i, n_{-i}'), (n_i, n_{-i})} = \WW_{(n_i', n_{-i}'), (n_i', n_{-i})} = \theta_\set{S}, 
  \end{align*}
  Which completes the proof. 
\end{proof}

\subsection{Proof of Theorem \ref{thm:equivar}}
\begin{proof}
  We will prove both the forward and backward direction: \\
  $( \Leftarrow)$ Suppose $\WW$ has the form given by (\ref{eqn:params_tensor}). We must show the layer is only equivariant with respect to permutations in $ \grn{S}{N_1} \times ... \times \grn{S}{N_D}$: 
  \begin{itemize}
    \item \textbf{Equivariance}: Let $\grn{g}{\pN} \in \grn{S}{N_1} \times ... \times \grn{S}{N_D}$, and let $\mat{G}\prm{\pN}$ be the corresponding permutation matrix. Then a simple extension of Theorem 2.1 in \citep{ravanbakhsh_symmetry} implies $\mat{G}\prm{\pN} \WW \XX = \WW \mat{G}\prm{\pN} \XX$ for all $\XX$, and thus the layer is equivariant. Intuitively, if $\grn{g}{\pN} \in \grn{S}{N_1} \times ... \times \grn{S}{N_D}$ we can ``decompose''  $\grn{g}{\pN}(\bn)$ into $D$ permutations $\grn{S}{N_1}(n_1),  ..., \grn{S}{N_D}(n_D)$ which act independently on the $D$ dimensions of $\XX$. 
    \item \textbf{No equivariance wrt any other permutation}: Let $\grn{g}{\pN} \in \grn{S}{\pN}$, with $\grn{g}{\pN} \not\in \grn{S}{N_1} \times \dots \times \grn{S}{N_D}$, and let $\mat{G}\prm{\pN}$ be the corresponding permutation matrix.
    Using Propositions \ref{prop:equivar1} and \ref{prop:equivar2} we have:
   \begin{align*}
     \mat{G}\prm{\pN} \WW \neq \WW \mat{G}\prm{\pN}
   \end{align*}
  So there exists an index at which these two matrices differ, call it $(\bn, \bn')$. Then if we take $\vec{\XX}$ to be the vector of all 0's with a single 1 in position $\bn'$, we will have:
    \begin{align*}
     \mat{G}\prm{\pN} \WW \vec{\XX} \neq \WW \mat{G}\prm{\pN} \vec{\XX} .
   \end{align*} 
   And since $\sigma$ is assumed to be strictly monotonic, we have:
   \begin{align*}
     \sigma(\mat{G}\prm{\pN} \WW \vec{\XX}) \neq \sigma( \WW \mat{G}\prm{\pN} \vec{\XX}) .
   \end{align*}  
   And finally, since \mat{G}\prm{\pN} is a permutation matrix and $\sigma$ is applied element-wise, we have:
   \begin{align*}
     \mat{G}\prm{\pN}\sigma( \WW \vec{\XX}) \neq \sigma( \WW \mat{G}\prm{\pN} \vec{\XX}) .
   \end{align*}     
   Therefore, the layer $\sigma( \WW \vec{\XX})$ is not a exchangeable tensor layer, and the proof is completed.
   \end{itemize}
  This proves the first direction. \\
  \\
  $( \Rightarrow)$ We prove the contrapositive.  Suppose $\WW_{\bn,\bn'} \neq \WW_{\bm,\bm'}$ for some $\bn,\bn',\bm,\bm' \in N_1 \times ... \times N_D$ with $\{ i \st n_i = n_i'\} = \{ i \st m_i = m_i'\}$. We want to show that the layer ${\YY} = \unvec{\sigma( \WW \vec{\XX} )}$ is not equivariant to some permutation $\grn{g}{\pN} \in \grn{S}{N_1} \times ... \times \grn{S}{N_D}$. We define this permutation as follows: 
   \begin{align*}
    \grn{g}{\pN}(\nu) = 
      \begin{cases}
        \bm \quad &\text{if} \quad \nu = \bn \\
        \bm' \quad &\text{if} \quad \nu = \bn' \\
        \bn \quad &\text{if} \quad \nu = \bm \\
        \bn' \quad &\text{if} \quad \nu = \bm' \\
        \nu  \quad &\text{otherwise}
      \end{cases}      
  \end{align*}
  That is, $\grn{g}{\pN}$ ``swaps'' $\bn$ with $\bm$ and $\bn'$ with $\bm'$. This is a valid permutation first since it acts element-wise, but also since $\{ i \st n_i = n_i'\} = \{ i \st m_i = m_i'\}$ implies that $\bn = \bn'$ iff $\bm = \bm'$ (and so $\grn{g}{\pN}$ is injective, and thus bijective). So if $\mat{G}\prm{\pN}$ is the permutation matrix of $\grn{g}{\pN}$ then we have $(\mat{G}\prm{\pN})_{(\bn',\bn)} = (\mat{G}\prm{\pN})_{(\bm',\bm)} = 1$, and:
  \begin{align*}
    (\mat{G}\prm{\pN} \WW)_{(\bm,\bn')} &= \sum_{\bi \in [N_1 \times ... \times N_D]} (\mat{G}\prm{\pN})_{(\bm,\bi)} \WW_{(\bi,\bn')} \\
    &= \WW_{(\bn,\bn')} \\
    &\neq \WW_{(\bm,\bm')} \\    
    &= \sum_{\bi \in [N_1 \times ... \times N_D]} \WW_{(\bm, \bi)} (\mat{G}\prm{\pN})_{(\bi, \bn')} \\
    &= (\WW \mat{G}\prm{\pN})_{(\bm,\bn')}
  \end{align*}

  And so $\mat{G}\prm{\pN} \WW \neq \WW \mat{G}\prm{\pN}$ and by an argument identical to the above, with appropriate choice of $X$, this layer does not satisfy the requirements of an exchangeable tensor layer. This completes the proof. 
\end{proof}

\subsection{Proof of Theorem \ref{thm:pe_layer}}
A simple reparameterization allows us to write the matrix $\WW$ of (\ref{eqn:params_tensor}) in the form of (\ref{eq:layer_simple_mat}). Thus Theorem \ref{thm:pe_layer} is just the special case of Theorem \ref{thm:equivar} where $D=2$ and the proof follows from that. 

\section{Details of Architecture and Training}

\textbf{\textsc{Self-Supervised Model.}}
Details of architecture and training: we train a simple feed-forward network with 9 exchangeable matrix layers using a leaky ReLU activation function. Each hidden layer has 256 channels and we apply a channel-wise dropout with probability 0.5 after the first to seventh layers. We found this channel-wise dropout to be crucial to achieving good performance. Before the input layer we mask out a proportion of the ratings be setting their values to 0 uniformly at random with probability 0.15. We convert the input ratings to one-hot vectors and interpret the model output as a probability distribution over potential rating levels. We tuned hyper-parameters by training on 75\% of the data, evaluating on a 5\% validation set. We test this model using the canonical u1.base/u1.test training/test split, which reserves 20\% of the ratings for testing.
For the MovieLens-1M dataset, we use the same architecture as for ML-100k and trained on 85\% of the data, validating on 5\%, and reserving 10\% for testing. The limited size of GPU memory becomes an issue for this larger dataset, so we had to employ conditional sampling for training. 
At validation time we used full batch predictions using the CPU in order to avoid memory issues.

\textbf{\textsc{Factorized Exchangeable Autoencoder Model.}}
Details of architecture and training: we use three exchangeable matrix layers for the encoder. The first two have 220 channels, and the third layer maps the input to 100 features for each entry, with no activation function applied. This is followed by mean pooling along both dimensions of the input. Thus, each user and movie is encoded into a length 100 vector of real-valued latent features. The decoder uses five similar exchangeable matrix layers, with the final layer having five channels. We apply a channel-wise dropout with probability 0.5 after the third and fourth layers, which we again found to be crucial for good performance. We convert the input ratings to one-hot vectors and optimize using cross-entropy loss.

\end{document}